# Multilingual Sentiment Analysis of Summarized Texts: A Cross-Language Study of Text Shortening Effects

Mikhail Krasitskii[1][†], Grigori Sidorov[1][†], Olga Kolesnikova[1*][†], Liliana Chanona Hernandez[2][†], Alexander Gelbukh[1][†]

[1*]Centro de Investigación en Computación (CIC), Instituto Politécnico Nacional (IPN), Av. Juan de Dios Bátiz S/N, Nueva Industrial Vallejo, Gustavo A. Madero, Ciudad de México, 07700, CDMX, México.
[2]Escuela Superior de Ingeniería Mecánica y Eléctrica (ESIME), Instituto Politécnico Nacional (IPN), Av. Luis Enrique Erro, Adolfo López Mateos S/N, Zacatenco, Gustavo A. Madero, Ciudad de México, 07700, CDMX, México.

*Corresponding author(s). E-mail(s): kolesnikova@cic.ipn.mx;
Contributing authors: mkrasitskii2023@cic.ipn.mx; sidorov@cic.ipn.mx; lchanona@gmail.com; gelbukh@cic.ipn.mx;
[†]These authors contributed equally to this work.

**Abstract**

Summarization significantly impacts sentiment analysis across languages with diverse morphologies. This study examines extractive and abstractive summarization effects on sentiment classification in English, German, French, Spanish, Italian, Finnish, Hungarian, and Arabic. We assess sentiment shifts post-summarization using multilingual transformers (mBERT, XLM-RoBERTa, T5, and BART) and language-specific models (FinBERT, AraBERT). Results show extractive summarization better preserves sentiment, especially in morphologically complex languages, while abstractive summarization improves readability but introduces sentiment distortion, affecting sentiment accuracy. Languages with rich inflectional morphology, such as Finnish, Hungarian, and Arabic, experience greater accuracy drops than English or German. Findings emphasize the need for language-specific adaptations in sentiment analysis and propose a hybrid summarization approach balancing readability and sentiment preservation. These insights benefit multilingual sentiment applications, including social media monitoring, market analysis, and cross-lingual opinion mining.

**Keywords:** Summarization, Sentiment analysis, Multilingual, Transformer

# Introduction

## Problem Statement

With the exponential growth of multilingual digital content, the need for efficient text processing techniques has become crucial for sentiment and opinion analysis. Sentiment analysis, a core task in Natural Language Processing (NLP), evaluates the sentiment of texts and is widely applied in domains such as social media monitoring, customer feedback analysis, and market trend prediction [1, 2].

Despite significant advancements in the development of multilingual natural language processing (NLP) systems, the effectiveness of these models continues to vary considerably across different languages. This inconsistency is primarily attributed to variations in morphological and syntactic structures. For instance, English, which is characterized by a relatively simple grammatical framework and a consistent word order, is more easily processed by automated systems. In contrast, languages such as Finnish, with its agglutinative features, and Arabic, known for its intricate inflectional morphology, present notable challenges. These difficulties become particularly evident when techniques for automated text summarization are employed [3, 4].

## Impact of Text Summarization on Sentiment Analysis

Text summarization reduces the length of a text while preserving key information; however, its impact on sentiment analysis varies depending on the summarization method used. Currently, there are two main types of text summarization: extractive (which retains key sentences from the original text) and abstractive (which paraphrases content using generative models). These approaches have different effects on the accuracy of sentiment analysis [3, 5].

**Extractive summarization** generally maintains sentiment accuracy, particularly in languages with fixed word order and simple morphology, such as English and German. Studies on sentiment classification using datasets like IMDb and GermEval show only a minor decrease in accuracy (**3–5%**) after extractive summarization [6, 7]. However, in morphologically rich languages, extractive methods may fragment meaning, reducing sentiment consistency.

**Abstractive summarization**, implemented using models such as T5 and BART, introduces sentiment distortion, especially in languages with complex morphology, including Finnish, Hungarian, and Arabic. Research indicates that summarization can reduce sentiment classification accuracy by more than **10%** in these languages [3, 8]. In Arabic, omitting inflectional markers alters sentiment perception [9, 10].

The impact on sentiment also varies depending on linguistic and cultural factors. Languages with flexible syntax, such as Finnish and Hungarian, exhibit greater sentiment distortion because abstractive summarization significantly alters sentence structure [11]. In contrast, European languages that explicitly express sentiment maintain higher accuracy after summarization [12].

Empirical evidence suggests that hybrid summarization approaches, combining extractive and controlled abstractive techniques, can more effectively balance readability and sentiment preservation [13, 14]. These findings highlight the need for language-specific adaptations in summarization for sentiment-sensitive tasks, such as social media monitoring and market research [15].

## Research Objective and Questions

This study investigates the impact of text summarization on sentiment analysis accuracy across multiple languages with varying morphological complexities and syntactical structures. The research focuses on eight languages English, German, French, Spanish, Italian, Finnish, Hungarian, and Arabic to evaluate how sentiment consistency is influenced by text shortening.

To achieve this, two summarization techniques extractive and abstractive are applied. Extractive summarization selects key sentences based on statistical and linguistic features, while abstractive summarization generates reformulated versions of the original content. The effects of these approaches on sentiment analysis are assessed using both multilingual and language-specific models. Among the multilingual models, mBERT and XLM-RoBERTa are employed for their broad applicability. Additionally, language-specific models such as FinBERT for Finnish and AraBERT for Arabic are utilized to enhance sentiment classification performance in morphologically rich languages.

For empirical validation, sentiment analysis is performed on datasets including IMDb for English, GermEval for German, SemEval for French, Spanish, and Italian, the Finnish Sentiment Dataset, the Hungarian Emotion Corpus, and ArSAS for Arabic. These datasets ensure balanced sentiment distribution and linguistic diversity.

To systematically analyze the effects of summarization on sentiment analysis, the following research questions are formulated:

**RQ1:** How does text summarization affect sentiment classification accuracy in languages with different morphological complexities?

**RQ2:** Which summarization approach (extractive or abstractive) effectively influences the preservation of text sentiment across different languages?

The results of this study are expected to provide insights into the relationship between summarization and sentiment classification across languages. Practical recommendations for optimizing sentiment analysis in summarized texts are formulated, benefiting applications in social media monitoring, market analysis, and multilingual NLP systems.

# Related Work

## Recent Developments

Automatic text summarization (ATS) methods have been developed since the mid-20th century and are categorized into abstractive and extractive approaches [16]. In recent years, significant progress in abstractive summarization has been achieved by applying neural network models, including transformer-based architectures such as T5, BART, and mT5 [14, 15]. These models generate coherent and contextually aware summaries, incorporating textual fragments that may not appear in the original document. While this approach enhances readability, it often alters the sentiment of the text, making it less suitable for sentiment analysis tasks [16, 17].

In contrast, extractive methods focus on selecting the most relevant sentences from the original text, ensuring the preservation of its structure and tonal integrity [16, 18]. Graph-based methods such as TextRank and LexRank analyze the internal structural relationships between sentences, contributing to high-quality information retention. However, despite their effectiveness, such approaches may overlook contextual elements essential for conveying emotional meaning [16]. Additionally, strict adherence to the original wording may sometimes reduce the fluency of the final summary.

Recent studies have also highlighted the importance of addressing data imbalance in sentiment analysis, particularly when dealing with multilingual and summarized texts. For instance, [?] proposed a method combining data balancing techniques, such as under-sampling and data augmentation through translation and paraphrasing, with transformer-based models like XLM-RoBERTa. Their approach significantly improved sentiment classification accuracy, especially for minority sentiment classes, achieving notable gains in precision, recall, and F1-score. This underscores the necessity of addressing data imbalance when applying sentiment analysis to multilingual and summarized texts, where sentiment distribution can vary significantly across languages and text lengths.

Moreover, the use of language-specific models, such as FinBERT for Finnish and AraBERT for Arabic, has further enhanced sentiment analysis accuracy in morphologically rich languages [6]. These models, pre-trained on language-specific corpora, capture linguistic intricacies that general multilingual models might overlook, thereby improving sentiment preservation in summarized texts.

## Challenges in Summarization for Sentiment Analysis

The process of text summarization is widely utilized for handling large volumes of textual data, particularly in domains such as social media, marketing, and healthcare. Two primary approaches extractive and abstractive summarization are commonly employed, each with inherent advantages and limitations.

One of the key challenges in summarization for sentiment analysis is the preservation of sentiment during text compression. [?] provides a comprehensive review of automatic text summarization methods, highlighting that extractive approaches, which preserve the original text structure, are often more effective in maintaining sentiment accuracy, particularly in morphologically rich languages. This is crucial for our study, as we investigate the impact of text shortening on sentiment preservation across multiple languages.

Furthermore, [?] proposes a method for multi-document summarization based on the relevance of linguistic and statistical features, such as thematic words, keywords, and POS tags. Their approach, which utilizes genetic algorithms for sentence selection optimization, demonstrates that the careful selection of features can significantly improve summarization quality and, consequently, sentiment preservation. These findings are particularly relevant to our research, as we aim to understand how summarization techniques affect sentiment accuracy in diverse linguistic contexts.

Extractive methods, which retain original phrases and sentence structures, often struggle with coherence, particularly in languages characterized by flexible word order, such as Finnish and Hungarian [19]. In contrast, abstractive summarization enables text reformulation while preserving core meaning but is prone to sentiment distortion. Research has demonstrated that models such as T5 and BART generate grammatically sound summaries yet frequently alter the sentiment, especially in morphologically rich languages where subtle word structure variations influence sentiment interpretation [7].

A major challenge in summarization for sentiment analysis is the preservation of sentiment in compressed texts. Studies indicate that abstractive summarization may inadvertently modify emotional perception, particularly in morphologically complex languages, where shifts in word order and syntactic structures significantly influence meaning [2]. Additionally, key challenges include:

- **Loss of Contextual Cues**: Sentiment-laden phrases are often removed during abstractive summarization, leading to alterations in meaning.
- **Structural Reformulation**: Languages with variable word order, such as Finnish and Hungarian, are subject to noticeable shifts in sentiment after summarization.
- **Semantic Drift**: Although the meaning may be preserved, sentiment shifts frequently result in sentiment polarity misclassification.

Moreover, standard summarization techniques frequently fail to account for cultural differences in sentiment interpretation. The complexity of cross-lingual sentiment analysis is further exacerbated in multilingual approaches, necessitating adaptations that integrate sociocultural factors.

Given these challenges, the necessity of hybrid summarization techniques that combine extractive and abstractive methods has been emphasized. Such approaches mitigate logical inconsistencies and reduce sentiment distortion, ultimately enhancing the accuracy and reliability of sentiment analysis in summarized texts [3].

## Language-Specific Insights

Research shows that summarization minimizes sentiment preservation in languages with relatively simple morphology, such as English and Spanish [5, 7]. However, in morphologically complex languages like Finnish and Arabic, significant changes in sentiment interpretation may occur. In particular, in Finnish, the removal of inflectional markers during text compression complicates sentiment analysis.

To improve classification accuracy, language-specific models such as FinBERT for Finnish and AraBERT for Arabic have been proposed, outperforming general multilingual models in sentiment preservation [3, 4]. This confirms the importance of tailored approaches to natural language processing.

The relationship between sentiment analysis and summarization also depends on linguistic structure. It has been found that in languages with a fixed word order, such as English and German, sentiment is better preserved than in languages with flexible syntax, such as Finnish [11].

Furthermore, cultural factors have a significant impact. As noted in the study by [15], the sentiment of a text can vary considerably depending on cultural context, emphasizing the need for adaptive sentiment evaluation metrics.

For low-resource languages such as Hungarian and Arabic, the creation of specialized datasets has been proposed, significantly improving the quality of both summarization and sentiment analysis [13].

## Methodology

This section outlines the methodology applied in this study, covering data collection, preprocessing, summarization techniques, sentiment analysis models, and evaluation metrics. The approach ensures reproducibility and validity across multiple languages.

### Data Collection and Preprocessing

The study utilized large-scale multilingual datasets from European, Finno-Ugric, and Semitic language groups. These datasets were selected to encompass diverse textual sources, including reviews, social media posts, and news articles, ensuring balanced sentiment distribution for robust evaluation of summarization effects.

Table 1. presents the datasets used, including IMDb, GermEval, SemEval (French, Spanish, Italian), the Finnish Sentiment Dataset, the Hungarian Emotion Corpus, and ArSAS. These datasets highlight linguistic variations, such as the use of expressive sentiment in short Italian and Spanish phrases and complex morphological structures in Finnish.

| Item | Quantity |
|---|---|
| IMDb (English) | 50,000 |
| GermEval (German) | 15,000 |
| SemEval (French) | 12,000 |
| SemEval (Spanish) | 10,000 |
| SemEval (Italian) | 11,500 |
| Finnish Sentiment Dataset (Finnish) | 8,000 |
| Hungarian Emotion Corpus (Hungarian) | 7,000 |
| ArSAS (Arabic) | 15,000 |

**Table 1** Dataset statistics for various languages.

The preprocessing pipeline included:

- **Tokenization**: Standardized tokenization procedures were applied based on language-specific requirements;
- **Stopword Removal**: Language-specific stopwords were removed to enhance feature extraction;
- **Lemmatization**: Words were reduced to their base form to improve generalization in sentiment classification;
- **Text Normalization**: Case normalization and special character removal were applied to ensure consistency across datasets.

## Summarization Methods

Two primary text summarization methods were applied to evaluate their impact on sentiment analysis: extractive and abstractive summarization.

**Extractive Summarization**. This approach selected the most relevant sentences based on features such as term frequency-inverse document frequency (TF-IDF) and sentence position. Its effectiveness was particularly noted in languages with a relatively fixed word order, such as English and German [20, 21]. By preserving original sentence structures, the method maintained syntactical integrity and reduced information loss. However, its limitations became evident in languages with flexible syntax, including Finnish and Hungarian, where crucial contextual elements were often omitted. Extractive summarization was implemented using ranking-based algorithms, as described by [22].

The extractive summarization process involved:

- Sentence segmentation;
- Computation of term importance using TF-IDF scores;
- Ranking sentences based on importance scores;
- Selection of the top-ranked sentences for summary generation.

Extractive summarization was applied using ranking-based approaches described in prior research.

**Abstractive Summarization**. Unlike extractive methods, abstractive summarization relies on deep learning models to generate condensed versions of texts while preserving their core meaning. The application of transformer-based models, such as T5 and BART, demonstrated high fluency and readability [7, 19]. However, challenges arose in morphologically rich languages, such as Arabic and Finnish, where sentiment distortion was observed due to structural reformulation. The propensity for generating paraphrased content instead of directly selecting key sentences made this method vulnerable to shifts in sentiment polarity. For implementation, the Hugging Face Transformers library was employed to fine-tune summarization models on multilingual datasets.

It was implemented using pretrained transformer models such as:

- T5 (Text-To-Text Transfer Transformer);
- BART (Bidirectional and Auto-Regressive Transformers).

The implementation included:

- Training the summarization models on multilingual datasets;
- Fine-tuning the models on language-specific corpora;
- Generating new summarized versions of the texts.

While abstractive summarization improved readability and fluency, it introduced semantic drift, particularly in morphologically rich languages such as Finnish and Arabic.

**Selection of Models**. The choice of summarization models was guided by their proven efficiency in multilingual settings. T5 and BART were selected due to their strong performance in abstractive summarization tasks, as demonstrated in prior studies [23]. Their ability to generate coherent summaries was advantageous for European languages with moderate morphological complexity. Meanwhile, extractive summarization relied on TF-IDF-based methods, which have shown stability across different language structures [11].

**Methods of Sentiment Analysis**

In this study, both multilingual and language-specific models were utilized for sentiment analysis, considering the morphological and syntactical characteristics of different languages.

Multilingual models such as mBERT and XLM-RoBERTa were selected due to their broad applicability in NLP tasks and their ability to facilitate cross-lingual knowledge transfer. The mBERT model [24] was trained on a vast corpus of multilingual texts, enabling it to identify common linguistic patterns and effectively analyze sentiment in ambiguous contexts. XLM-RoBERTa [25] is an enhanced version of RoBERTa, optimized for multilingual text processing through deeper pretraining and an expanded training dataset.

For languages with high morphological complexity, specialized models were employed, including FinBERT [26] for Finnish and AraBERT [9] for Arabic. These models were pre-trained on language-specific corpora, taking into account agglutination and word formation processes. For instance, FinBERT incorporates adapted embeddings to account for Finnish suffixation, improving sentiment analysis accuracy. AraBERT, in turn, captures Arabic morphological characteristics, including root-based word structures and diacritical variations.

The selection of these models was justified by their effectiveness, as confirmed in previous studies [1, 3], as well as their ability to adapt to linguistic specificities, making them suitable for this research.

**Multilingual Sentiment Analysis Models**

Multilingual models were used to ensure broad applicability across multiple languages. The selected models included:

- mBERT (Multilingual Bidirectional Encoder Representations from Transformers): Pretrained on a diverse set of multilingual texts, enabling efficient cross-lingual sentiment classification;
- XLM-RoBERTa: An improved version of RoBERTa optimized for multilingual applications.

### Language-Specific Sentiment Models

For languages with high morphological complexity, specialized models were utilized:

- FinBERT for Finnish sentiment analysis;
- AraBERT for Arabic sentiment analysis.

These models were pretrained on language-specific corpora, capturing linguistic intricacies such as agglutination in Finnish and root-based morphology in Arabic.

## Evaluation Metrics

To evaluate the performance of models and methods, both traditional and adapted metrics were employed.

### Sentiment Classification

- Accuracy: Measures overall correctness of predictions;
- Precision: The percentage of correctly predicted positive/negative sentiments;
- Recall: The ability of the model to identify all relevant instances;
- F1-Score: Balances precision and recall, particularly crucial for imbalanced datasets.

### Summarization Quality

- ROUGE: Evaluates the overlap between summaries and original texts [27, 28];
- BLEU: Assesses grammaticality and fluency of generated abstractive texts [23].

### Semantic Similarity

Cosine similarity was calculated between embeddings of original and summarized texts to measure semantic loss, which is particularly relevant for morphologically rich languages [3, 7].

## Methodological Framework

The methodological framework employed in this study is designed to ensure scientific rigor, reproducibility, and scalability in multilingual sentiment analysis. It encompasses key components such as data collection, preprocessing techniques, summarization methods, sentiment analysis models, and evaluation metrics.

Table 2. provides a structured methodological framework. The data collection process integrates a diverse set of sentiment-labeled datasets across multiple languages, including IMDb, GermEval, SemEval, Finnish Sentiment Dataset, and ArSAS. Preprocessing steps such as tokenization, stopword removal, lemmatization, and text normalization were applied to standardize the text data.

The study employs two summarization techniques: extractive summarization (using TF-IDF and graph-based methods) and abstractive summarization (leveraging transformer-based models such as T5 and BART). For sentiment classification, multilingual models (mBERT, XLM-RoBERTa) and language-specific models (FinBERT for Finnish and AraBERT for Arabic) were utilized. Performance evaluation was conducted using multiple metrics, including accuracy, precision, recall, F1-score, ROUGE, BLEU, and cosine similarity.

| Methodological Component | Approach |
|---|---|
| Data Collection | IMDb, GermEval, SemEval, Finnish Sentiment Dataset, ArSAS |
| Preprocessing | Tokenization, stopword removal, lemmatization, text normalization |
| Summarization Techniques | Extractive (TF-IDF, Graph-based), Abstractive (T5, BART) |
| Sentiment Analysis Models | Multilingual: mBERT, XLM-RoBERTa; Language-specific: FinBERT, AraBERT |
| Evaluation Metrics | Accuracy, Precision, Recall, F1-Score, ROUGE, BLEU, Cosine Similarity |

**Table 2** Overview of Methodological Components

This structured methodology facilitates a comprehensive analysis of the impact of text summarization on sentiment classification across different linguistic systems. The following sections present the results and discussion derived from this framework.

# Results

## Overview of the Obtained Results

The study analyzed the effect of text summarization on the accuracy of sentiment analysis in various languages, taking into account their morphological and syntactic complexity. A comparison of sentiment classification results before and after summarization was conducted, including extractive and abstractive methods. The study aimed to determine the extent to which summarization affects sentiment preservation and classification effectiveness.

It was found that the impact of summarization techniques on sentiment classification varies depending on the language structure. In languages with less complex morphology, such as English, German, and Spanish, a slight decrease in classification accuracy was observed. At the same time, in languages with rich morphology and flexible syntax, such as Finnish, Hungarian, and Arabic, significant deviations in sentiment were recorded after summarization, which could be exacerbated when noisy or low-quality data were used.

Extractive summarization demonstrated higher effectiveness in preserving sentiment accuracy compared to abstractive summarization, especially in languages with complex morphology, due to the retention of original text structures. Abstractive summarization, despite improving text fluency, often leads to sentiment distortions, particularly in languages where word forms and their order are critically important.

To evaluate the results, metrics such as Accuracy, Precision, Recall, and F1-score were used to measure the effectiveness of sentiment classification, while ROUGE and BLEU were employed to assess summarization quality. A comparison of these metrics before and after summarization revealed the extent of sentiment changes caused by text shortening and confirmed the importance of data quality as a key factor influencing the final results.

Thus, it was demonstrated that to reduce distortions and improve the accuracy of sentiment analysis, it is necessary to consider both the morphological features of the language and the quality of the data used. The higher the purity and detail of the data, the more accurately sentiment is preserved during analysis.

## Sentiment Analysis before Summarization

The sentiment classification results before summarization indicated high accuracy levels across all languages. A baseline evaluation was conducted using multilingual models such as mBERT and XLM-RoBERTa, alongside language-specific models like FinBERT and AraBERT for Finnish and Arabic, respectively. The classification performance metrics for each language are presented in Table 3.

| Language | Precision (%) | Recall (%) | F1-Score (%) | Accuracy (%) |
|---|---|---|---|---|
| English | 90.4499 | 91.7499 | 91.0499 | 92.1499 |
| French | 87.9499 | 88.8499 | 88.3499 | 89.5499 |
| German | 86.7499 | 87.5499 | 87.1499 | 88.2499 |
| Spanish | 85.6499 | 86.2499 | 85.9499 | 87.0499 |
| Italian | 84.2499 | 85.0499 | 84.6499 | 86.3499 |
| Finnish | 80.9499 | 82.1499 | 81.5499 | 84.3499 |
| Hungarian | 79.8499 | 80.5499 | 80.2499 | 83.5499 |
| Arabic | 77.6499 | 78.9499 | 78.2499 | 81.7499 |

**Table 3** Sentiment Classification Metrics before Summarization

Before text summarization, sentiment classification metrics were obtained for multiple languages. The highest accuracy was achieved for English (92.1499%), followed by French (89.5499%) and German (88.2499%). In Spanish and Italian, accuracy values were 87.0499% and 86.3499%, respectively. For Finnish, Hungarian, and Arabic, accuracy was lower, ranging from 84.3499% to 81.7499%. A similar trend was observed for the Precision, Recall, and F1-score metrics, with English showing the best results and Arabic the lowest. Overall, higher metric values were recorded for languages of Latin and Germanic origin, while lower scores were noted for Finno-Ugric and Semitic languages.

## Summarization Results

The effects of summarization on sentiment classification were analyzed separately for extractive and abstractive methods. The quality of summarization was assessed using ROUGE-1 and BLEU metrics, which measure the overlap between the original and summarized texts, as well as the fluency and coherence of the generated summaries.

*Extractive Summarization Results

Extractive summarization, which selects key sentences from the original text, demonstrated higher consistency in preserving sentiment, especially in languages with complex morphology. The ROUGE-1 and BLEU scores for extractive summarization are presented in Table 4.

As shown in Table 4., English and French achieved the highest ROUGE-1 scores (89.3% and 88.5%, respectively), indicating a strong overlap between the original and summarized texts. Finnish and Arabic, on the other hand, showed lower scores (78.9% and 75.3%), reflecting greater content loss during summarization. Similarly, BLEU scores were highest for English (85.1%) and French (83.7%), while Arabic and Hungarian had the lowest scores (70.5% and 72.3%), indicating challenges in maintaining linguistic alignment and fluency.

| Language | ROUGE-1 (%) |
|---|---|
| English | 89.3 |
| French | 88.5 |
| German | 86.8 |
| Spanish | 86.2 |
| Italian | 85.4 |
| Finnish | 78.9 |
| Hungarian | 76.7 |
| Arabic | 75.3 |

**Table 4** Extractive Summarization Results (ROUGE-1 Scores).

**\*Abstractive Summarization Results**

Abstractive summarization, which generates paraphrased versions of the original text, introduced greater variability in sentiment preservation. The ROUGE-1 and BLEU scores for abstractive summarization are presented in Table 5

| Language | ROUGE-1 (%) | BLEU (%) |
|---|---|---|
| English | 87.1 | 83.2 |
| French | 86.3 | 82.1 |
| German | 84.7 | 80.9 |
| Spanish | 83.9 | 79.8 |
| Italian | 82.5 | 78.6 |
| Finnish | 74.8 | 70.2 |
| Hungarian | 72.4 | 68.7 |
| Arabic | 70.1 | 66.3 |

**Table 5** Abstractive Summarization Results (ROUGE-1 and BLEU Scores)

As shown in Table 5., abstractive summarization resulted in lower ROUGE-1 and BLEU scores compared to extractive summarization across all languages. English and French still achieved relatively high scores (87.1% and 86.3% for ROUGE-1, and 83.2% and 82.1% for BLEU), but the drop was more pronounced in morphologically complex languages like Finnish, Hungarian, and Arabic. For example, Arabic had the lowest ROUGE-1 score (70.1%) and BLEU score (66.3%), indicating significant challenges in maintaining both content overlap and fluency.

Figure 1. presents a comparative analysis of ROUGE-1 and BLEU scores for extractive and abstractive summarization across different languages.

As shown by the **ROUGE-1** metric (Figure 1.), extractive summarization consistently outperforms abstractive summarization in terms of content preservation, with the most significant differences observed for Finnish, Hungarian, and Arabic.

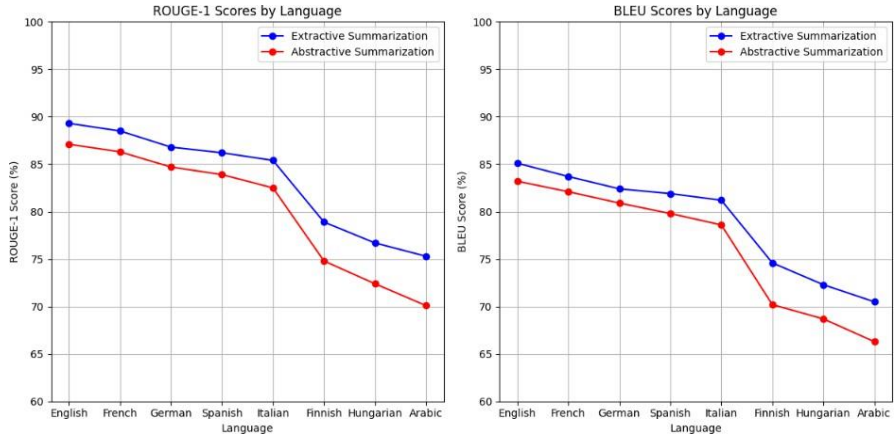

**Fig. 1** Comparison of ROUGE-1 and BLEU Scores for Extractive and Abstractive Summarization by Language

A similar trend was observed in the **BLEU** metric (Figure 1.), where extractive summarization ensured higher coherence and fluency, particularly in languages with simpler morphology, such as English and French. Abstractive summarization, while improving readability, struggled to maintain linguistic alignment in morphologically rich languages.

## Sentiment Analysis after Summarization

The sentiment classification performance following summarization was assessed using the same evaluation metrics as before summarization. The results are presented in Table 6.

| Language  | Precision (%) | Recall (%) | F1-Score (%) | Accuracy (%) |
|-----------|---------------|------------|--------------|--------------|
| English   | 87.8499       | 88.6499    | 88.2499      | 89.5499      |
| French    | 84.7499       | 85.4499    | 85.0499      | 86.2499      |
| German    | 83.5499       | 84.2499    | 83.8499      | 85.6499      |
| Spanish   | 81.6499       | 82.2499    | 81.9499      | 83.5499      |
| Italian   | 80.5499       | 81.2499    | 80.8499      | 82.7499      |
| Finnish   | 73.8499       | 74.9499    | 74.3499      | 77.4499      |
| Hungarian | 71.7499       | 72.5499    | 72.1499      | 76.0499      |
| Arabic    | 71.2499       | 72.4499    | 71.8499      | 75.6499      |

**Table 6** Sentiment Classification Metrics after Summarization

The sentiment classification results after summarization demonstrated variations across languages. The highest accuracy was recorded for English (89.5499%), followed by French (86.2499%) and German (85.6499%). For the remaining languages, accuracy ranged from 83.5499% (Spanish) to 75.6499% (Arabic). Precision, recall, and F1-score followed a similar trend, with the best performance observed for English (88.2499%

F1-score) and the lowest for Arabic (71.8499%). Overall, higher metric values were characteristic of languages using the Latin alphabet, while Finnish, Hungarian, and Arabic exhibited lower results.

## Comparison of Sentiment Metrics before and after Summarization

The analysis showed that text summarization negatively affects sentiment classification across all languages. The most significant decline in accuracy, recall, precision, and F1-score is observed in Finnish, Hungarian, and Arabic, which can be attributed to their morphological complexity. Meanwhile, in languages with simpler morphology, such as English and French, the impact of summarization is less pronounced.

A comparison of different summarization methods revealed that sentiment-associated information is better preserved by extractive approaches than by abstractive ones.

As shown in Figure 2., classification **accuracy** decreases in all languages, with the most substantial drop in Finnish, Hungarian, and Arabic. A similar trend is observed for **precision** (Figure 3.), **recall** (Figure 4.), and **F1-score** (Figure 5.), where the most significant performance degradation is recorded in morphologically complex languages.

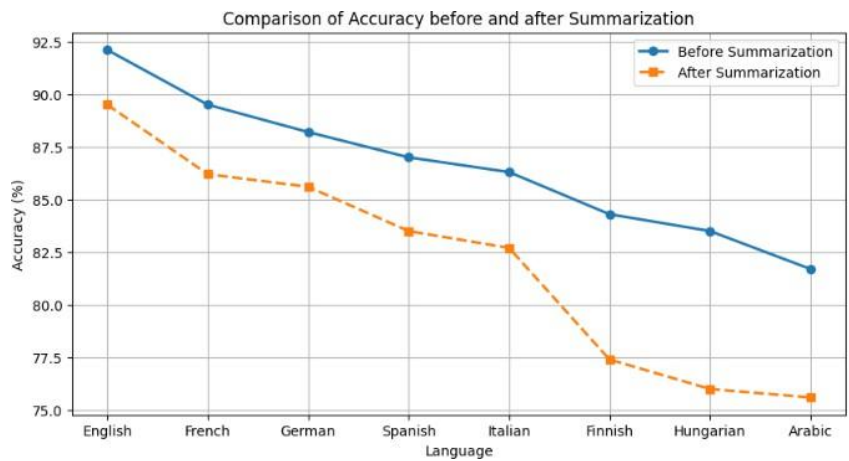

**Fig. 2** Comparison of Accuracy before and after Summarization

## Summary of Findings

The study reveals that text summarization significantly impacts sentiment analysis accuracy, with the extent of the effect varying by language morphology. Extractive summarization consistently outperforms abstractive methods in preserving sentiment, particularly in morphologically complex languages like Finnish, Hungarian, and Arabic. While improving readability, Abstractive summarization introduces greater sentiment distortion, especially in languages where word order and inflectional morphology are critical.

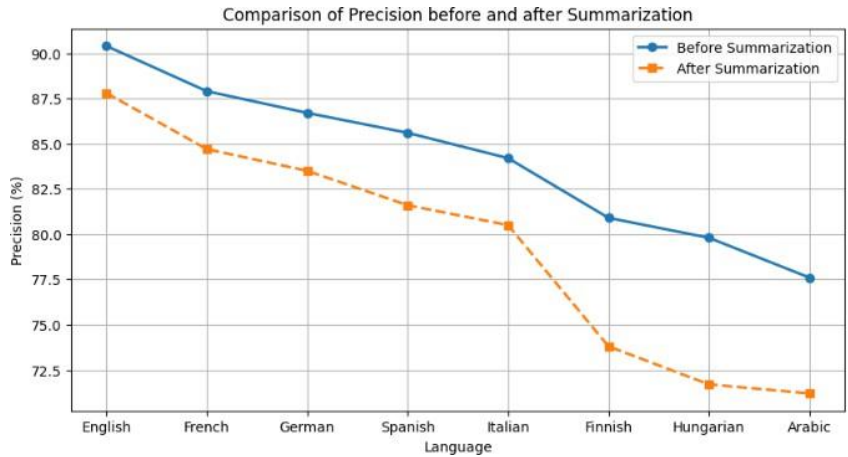

**Fig. 3** Comparison of Precision before and after Summarization

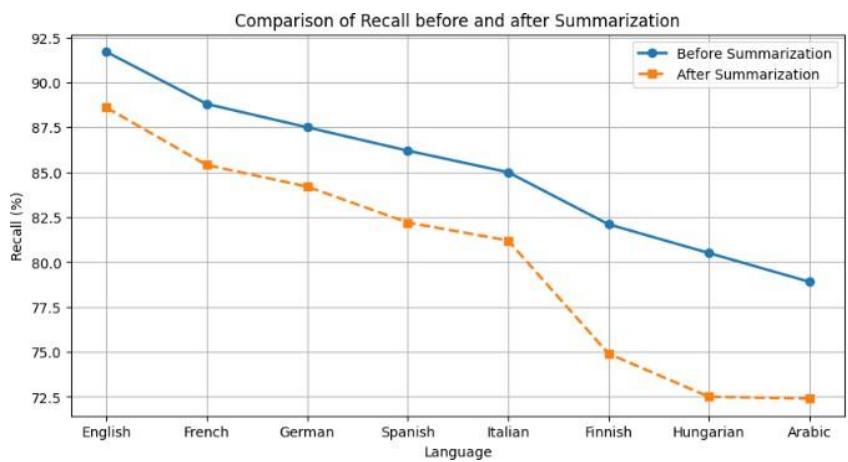

**Fig. 4** Comparison of Recall before and after Summarization

Languages with simpler morphology, such as English and German, experience minimal accuracy drops after summarization. In contrast, morphologically rich languages show significant declines in sentiment classification accuracy, exacerbated by noisy or low-quality data. Extractive methods that retain original sentence structures maintain higher sentiment consistency, whereas abstract methods often lead to semantic drift and sentiment polarity misclassification.

Evaluation metrics, including ROUGE-1 and BLEU, confirm that extractive summarization better preserves content overlap and fluency. As illustrated in Figure 6., the percentage changes in accuracy across languages highlight the superior performance of extractive summarization in preserving sentiment, particularly in complex languages like Finnish and Arabic. These findings underscore the importance of language-specific summarization strategies, balancing readability and sentiment preservation, especially in multilingual applications such as social media monitoring and market analysis.

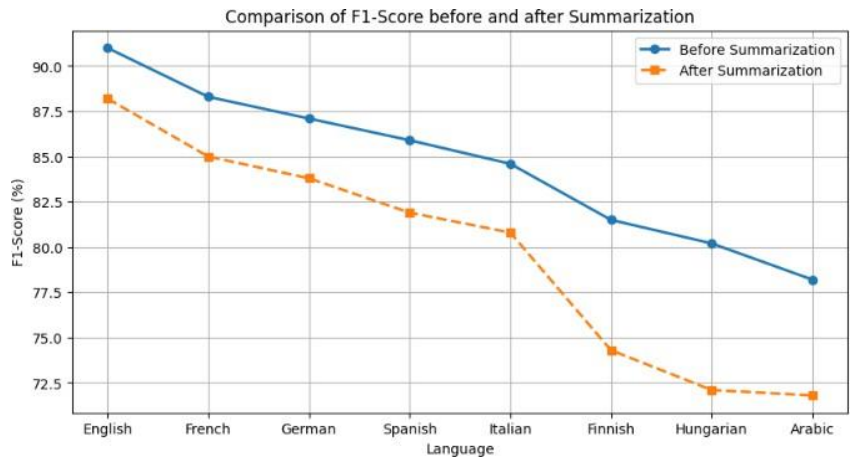

**Fig. 5** Comparison of F1-Score before and after Summarization

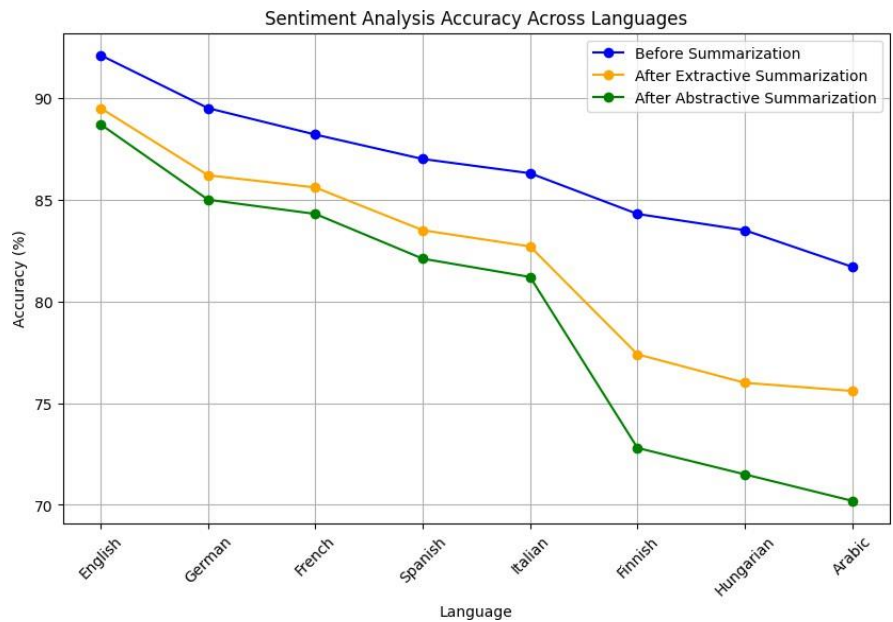

**Fig. 6** Sentiment Analysis Accuracy Across Languages

# Discussion

The findings of this study emphasize the significant impact of text summarization on sentiment analysis across languages with diverse morphological and syntactical structures. The results demonstrate that while summarization enhances text readability and reduces processing time, it also introduces variability in sentiment classification accuracy, particularly in morphologically rich languages. This discussion examines the implications of these findings, compares them with prior research, and proposes potential improvements for sentiment analysis models.

## Impact of Summarization on Sentiment Classification

The study reveals that the extent of sentiment alteration due to summarization varies across languages. In languages with relatively simple morphology, such as English and Spanish, the effect of summarization is minimal. However, in languages with complex morphological structures, such as Finnish and Arabic, summarization results in a significant drop in sentiment classification accuracy.

Extractive summarization, which preserves the most relevant sentences from the original text, maintains sentiment accuracy more effectively than abstractive summarization. The latter, which generates restructured versions of the original content, often leads to sentiment distortion, particularly in morphologically complex languages. This supports findings from previous research indicating that abstractive summarization may introduce sentiment shifts due to its paraphrasing nature.

## Language-Specific Models vs. Multilingual Approaches

A notable observation from the study is the superiority of language-specific models (e.g., FinBERT for Finnish and AraBERT for Arabic) in maintaining sentiment accuracy compared to general multilingual models. These specialized models, trained on native-language corpora, outperform general-purpose transformers in languages with high morphological complexity.

However, multilingual models remain advantageous for low-resource languages, where language-specific models are unavailable. A hybrid approach that integrates multilingual models with fine-tuning on language-specific datasets could enhance overall sentiment classification performance.

## Challenges in Maintaining Sentiment Consistency

The study identifies key challenges in sentiment preservation during summarization:

- Loss of Contextual Cues: Abstractive summarization often removes essential sentiment-bearing words or phrases, leading to misclassification.
- Structural Reformulation: Sentence restructuring in summarization can alter sentiment expression, especially in languages where syntax plays a crucial role in sentiment interpretation.
- Semantically Equivalent but Emotionally Divergent Paraphrasing: Rewriting a sentence with different phrasing may preserve meaning but alter the emotional connotation.

These challenges necessitate improved summarization models that incorporate sentiment retention mechanisms, particularly for applications in social media monitoring, customer feedback analysis, and multilingual opinion mining.

## Potential Solutions and Future Directions

To mitigate the sentiment distortions introduced by summarization, several strategies are proposed:

- Hybrid Summarization Approaches: Combining extractive and abstractive summarization techniques could balance fluency and sentiment preservation.
- Fine-Tuning for Sentiment Sensitivity: Training summarization models specifically for sentiment analysis tasks may enhance accuracy.
- Leveraging Contextual Embeddings: Integrating contextual embeddings from language models that preserve sentiment polarity could reduce sentiment drift.
- Cultural Adaptation in Sentiment Models: Considering cultural nuances in sentiment interpretation can improve multilingual sentiment analysis performance.

# Conclusions and Recommendations

## Conclusion

This study highlights that the accuracy of sentiment analysis in multilingual contexts is significantly influenced by text summarization, with the extent of this impact largely determined by the morphological complexity of the language and the quality of the analyzed data. The key findings can be summarized as follows:

The **morphological complexity** of a language is one of the key factors influencing how text summarization affects sentiment analysis. Languages with relatively simple morphology, such as English, German, and Spanish, show minimal changes in sentiment after summarization. In contrast, languages with rich and complex morphology, such as Finnish, Hungarian, and Arabic, are more prone to noticeable deviations in sentiment. This is due to the specificities of word formation and syntactic flexibility in these languages, which can be disrupted during the summarization process.

Furthermore, the **quality of the data** also plays a crucial role. High-quality and detailed data tend to preserve sentiment more accurately even after summarization. At the same time, noisy or low-quality data can exacerbate sentiment distortions, especially in languages with complex morphology. This underscores the need for thorough data preprocessing and selection to ensure reliable sentiment analysis results.

**Extractive summarization**, which preserves the original sentence structure, proves to be more effective in maintaining sentiment accuracy compared to abstractive methods. This is particularly relevant for languages with complex morphology, where preserving the original syntactic and morphological structure is critical for the correct interpretation of sentiment.

**Abstractive summarization**, while improving text fluency, carries a higher risk of sentiment distortion. This is especially evident in languages where word order and morphological changes play a key role in conveying sentiment. Poor data quality can exacerbate these distortions, making abstractive methods less reliable for sentiment-sensitive tasks in such languages.

**Language-specific sentiment analysis models**, such as FinBERT for Finnish and AraBERT for Arabic, outperform general multilingual models in preserving sentiment accuracy after text summarization. These models, trained on high-quality, language-specific data, are better equipped to handle the unique morphological and syntactic features of their respective languages, leading to more accurate sentiment analysis.

Thus, the study's results emphasize that both the morphological characteristics of a language and the quality of the data are critical factors influencing the accuracy of sentiment analysis after text summarization. Effective management of these factors is necessary to minimize sentiment distortions and achieve reliable results in multilingual sentiment analysis tasks.

## Recommendations

To minimize sentiment distortion during text summarization, the following recommendations are proposed:

**Use of Hybrid Summarization Strategies**: Retaining key phrases containing emotionally significant elements through extractive methods. Controlled use of abstractive summarization to improve readability while preserving sentiment integrity.

**Adaptation of Language Models for Sentiment Analysis**: Utilizing specialized language models (e.g., FinBERT for Finnish and AraBERT for Arabic) to improve sentiment classification accuracy in morphologically complex languages. Fine-tuning multilingual models on sentiment-specific corpora to reduce emotional information loss.

**Optimization of NLP Models for Sentiment Preservation**: Training summarization models with a focus on sentiment-related characteristics. Integrating attention mechanisms capable of identifying words and phrases critical to conveying sentiment.

**Development of Sentiment-Oriented Evaluation Metrics**: Refining standard metrics (ROUGE, BLEU) to incorporate sentiment characteristics. Introducing new metrics capable of assessing sentiment consistency between original and summarized texts.

**Consideration of Cultural and Contextual Sentiment Factors**: Enhancing sentiment analysis models with sociocultural adaptations to account for variations in emotion perception across languages. Applying specialized models for analyzing reviews and social media in multilingual environments.

**Utilization of Next-Generation Transformer Models for Multilingual Sentiment Analysis**: Exploring the potential of advanced models such as LLAMA 2 and BLOOM for improving sentiment analysis in summarized texts. Conducting further research into their applicability for summarization tasks with a focus on sentiment preservation.

**Application of Improved Summarization Methods in Real-World Scenarios**: Integrating adapted methods into review analysis systems, social media monitoring, and market research. Optimizing existing solutions for multilingual settings, considering language-specific sentiment transmission.

These recommendations aim to enhance sentiment analysis accuracy in summarized texts and minimize distortions occurring during summarization across different languages. By adopting these strategies, researchers and practitioners can improve the reliability and effectiveness of sentiment analysis in multilingual contexts.

## Declaration

### Author Contributions

M.K., and O.K. played pivotal roles in the experimental design and data collection, while M.K. spearheaded the data analysis and interpretation. The initial manuscript was drafted by M.K. and O.K., with revisions contributed by G.S., L.C.H. and A.G. All authors collectively approved the final manuscript. Notably, M.K., G.S., O.K., L.C.H., and A.G. equally share authorship and take joint responsibility for the accuracy and integrity of the entire work.

### Funding

The work was done with partial support from the Mexican Government through grant A1-S-47854 of CONAHCYT, Mexico, and grants 20241816, 20241819, and 20240951 of the Secretarıa de Investigacion y Posgrado of the Instituto Polit´ecnico Nacional, Mexico. The authors thank the CONAHCYT for the computing resources brought to them through the Plataforma de Aprendizaje Profundo para Tecnologıas del Lenguaje of the Laboratorio de Supercomputo of the INAOE, Mexico, and acknowledge the support of Microsoft through the Microsoft Latin America PhD Award. Additionally, we acknowledge the invaluable feedback and guidance provided by our peers during the review process. We are also grateful to the Instituto Polit´ecnico Nacional for providing the necessary infrastructure and resources to carry out this research. Finally, we extend our thanks to the developers of open-source tools and libraries, whose work significantly facilitated the technical aspects of our project.

### Data Availability

This manuscript does not report data generation.

### Conflict of Interest

The authors declare no competing interests